\documentclass[a4paper,11pt]{article}
\usepackage[margin=1in]{geometry} 
\usepackage{times}
\usepackage{latexsym}

\usepackage{xcolor} 
\usepackage{colortbl}

\usepackage[utf8]{inputenc} 
\usepackage[T1]{fontenc}    

\usepackage[numbers,compress]{natbib}

\usepackage[colorlinks,citecolor=blue]{hyperref}         
\usepackage{url}            
\usepackage{booktabs}       
\usepackage{amsfonts}       
\usepackage{nicefrac}       
\usepackage{microtype}      
\usepackage{graphicx}
\usepackage{todonotes}
\usepackage{amsmath}
\usepackage{amssymb}
\usepackage{amsthm}
\usepackage{array}
\usepackage{xcolor}
\usepackage{enumitem}
\usepackage{framed}
\usepackage{xspace}
\usepackage{booktabs}
\usepackage{enumitem}
\usepackage{hyperref}
\usepackage{multirow}
\usepackage{multicol}
\usepackage{tabularx}

\usepackage{xspace}
\usepackage{amsmath}
\usepackage{amsfonts}
\usepackage{mathtools}
\usepackage{xcolor,colortbl}
\usepackage{bm}
\usepackage{algorithm}
\usepackage{algorithmic}
\usepackage{multirow}
\usepackage{dsfont}
\usepackage{pict2e}

\usepackage{microtype}
\usepackage{graphicx}

\usepackage{amssymb}
\usepackage{amsthm}

\usepackage{subfig}
\usepackage{caption}
\usepackage{makecell}

\usepackage{amsmath}

\usepackage{booktabs}

\usepackage{enumitem}
\usepackage{comment}

\usepackage{csquotes}
\usepackage{vcell}
\usepackage{tabularx}
\usepackage[para]{footmisc}
\usepackage{tablefootnote}

\newcommand{\ie}{\textit{i.e.}\xspace}

\newcommand{\Wbf}{\mathbf{W}\xspace}

\newcommand{\Thetabf}{\bm{\Theta}\xspace}

\newcommand{\CE}{\mathcal{E}\xspace}

\newcommand{\CK}{\mathcal{K}\xspace}
\newcommand{\CM}{\mathcal{M}\xspace}

\newcommand{\CO}{\mathcal{O}\xspace}
\newcommand{\CP}{\mathcal{P}\xspace}

\definecolor{Gray}{gray}{0.95}
\newcolumntype{g}{>{\columncolor{Gray}}c}

\newtheorem{proposition}{Proposition}

\newcolumntype{P}[1]{>{\centering\arraybackslash}p{#1}}

\usepackage{etoolbox}

\renewenvironment{quote}
  {\begin{list}{}%
     {\setlength{\leftmargin}{5mm} 
      \setlength{\rightmargin}{5mm}} 
     \item\relax}
  {\end{list}}

\newcolumntype{H}{>{\setbox0=\hbox\bgroup}c<{\egroup}@{}}

\date{}

\title{Efficient Knowledge Probing of Large Language Models \\ by Adapting Pre-trained Embeddings}

\author{
    Kartik Sharma$^{1}$,Yiqiao Jin$^{1}$, Rakshit Trivedi$^{2}$, Srijan Kumar$^{1}$ \\
    $^{1}$Georgia Institute of Technology, 
    $^{2}$Massachusetts Institute of Technology\\
    $^{1}$\texttt{\{ksartik,yjin328,srijan\}@gatech.edu} 
    \\
    $^{2}$\texttt{triver@mit.edu} 
    \\
}

\begin{document}

\maketitle

\begin{abstract}
    Large language models (LLMs) acquire knowledge across diverse domains such as science, history, and geography encountered during generative pre-training.  
    However, due to their stochasticity, it is difficult to predict what LLMs have acquired. 
    Prior work has developed different ways to probe this knowledge by investigating the hidden representations, crafting specific task prompts, curating representative samples, and estimating their uncertainty. 
    However, these methods require making forward passes through the underlying model to probe the LLM's knowledge about a specific fact, making them computationally expensive and time-consuming.
    To bridge this gap, we propose \textbf{PEEK} or \textbf{P}roxy \textbf{E}mbeddings to \textbf{E}stimate \textbf{K}nowledge of LLMs, by leveraging the pre-trained embedding models that effectively encode factual knowledge as text or graphs as proxies for LLMs. 
    First, we identify a training set of facts known by LLMs through various probing strategies and then adapt embedding models to predict the LLM outputs with a linear decoder layer.
    Comprehensive evaluation on $3$ Wikipedia-derived datasets, $4$ LLMs, and $7$ embedding models shows that embeddings can predict LLM knowledge on a held-out set with up to 90\% accuracy.
    Furthermore, we find that sentence embedding models are more suitable than graph embeddings to predict LLM knowledge, shedding light on the underlying representation of factual landscape.
    Thus, we believe that knowledge-adapted embeddings can be used to identify knowledge gaps in LLMs at scale and can provide deeper insights into LLMs' internal inductive bias. 
    The code and data is made available at \url{https://github.com/claws-lab/peek}. 
\end{abstract}

\section{Introduction}\label{sec:introduction}

\begin{figure}[t]
    \centering
    \includegraphics[width=0.6\linewidth]{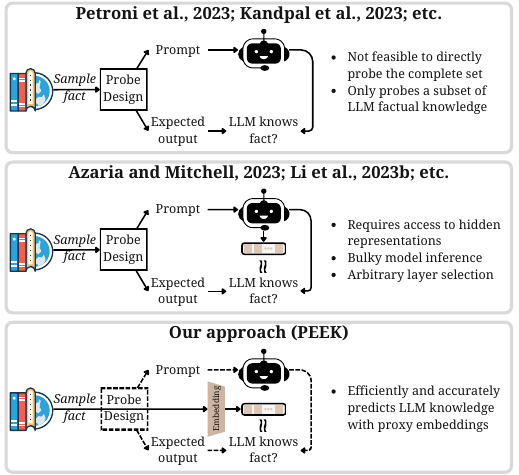}
    \caption{Comparison of our proposed approach, Proxy Embeddings to Estimate Knowledge (PEEK) with other knowledge probing approaches.}
    \label{fig:motivation}
\end{figure}

The internet serves as a modern-day encyclopedia and ever-expanding repository of human knowledge spanning history, science, technology, and arts~\citep{lehmann2015dbp100k,suchanek2007yago310}. 
Large language models (LLMs) trained on such web-scale data have thus become general-purpose knowledge bases~\citep{petroni2019language}, but it is still unclear to map their learned knowledge landscape due to stochastic learning objectives and noisy training data.
As LLMs are being adopted for knowledge-intensive applications such as healthcare~\citep{singhal2023large}, law~\citep{guha2023legalbench}, scientific discovery~\citep{shojaee2024llm,ma2024llm}, etc., it is essential to efficiently probe their acquired knowledge since hallucinating incorrect knowledge~\citep{huang2023hallucination} or lack of knowledge can diminish their utility. Thus, factual gaps are often identified and fixed before deployment through extensive risk assessment manually~\citep{shevlane2023model,khlaaf2022hazard}. 

Different techniques have been proposed in the literature to probe the knowledge of LLMs. This involves crafting specific knowledge probing prompts on a subsampled set of facts~\citep{petroni2019language,zheng2023kglens,kandpal2023large,sun2023head,luo2023systematic,jiang2020canknow}, investigating their hidden representations for non-factuality~\citep{azaria2023internal,gottesman2024estimating,wang2024faclens}, and estimating their epistemic uncertainty~\citep{ahdritz2024distinguishing,manakul2023selfcheckgpt,jesson2025estimating}. However, none of these methods can uncover the knowledge landscape of LLMs without making forward passes through the LLMs or peeking into their activations. This limits their applicability to extremely large and black-box language models. 

Concurrently, representation learning models have also become more powerful in learning general-purpose low-dimensional representations of real-world data such as multilingual texts~\citep{muennighoff2022mteb}, and knowledge graphs~\citep{galkin2023ultra}. These encoder models are optimized via self-supervised learning on curated subsets of web data, often overlapping with the training corpora of decoder-only LLMs. More recently, pre-trained LLMs have also been directly used to initialize these representations for various domains~\citep{behnamghader2024llm2vec,wang2024llms}. Due to the shared pre-training data and architectural similarities, these embeddings offer an efficient and accurate means of estimating the knowledge acquired by black-box LLMs.

In this work, we thus propose \textbf{PEEK}, where we use pre-trained embedding models to estimate the knowledge of an LLM for a set of training facts and then, using these embeddings as proxies to probe the knowledge of LLMs over a database. Figure~\ref{fig:motivation} illustrates our contributions in comparison to the literature. In particular,
\begin{enumerate}[leftmargin=*,itemsep=0mm]
    \item We propose a \textbf{novel approach} of probing LLM knowledge through proxy embedding models that are adapted using a trainable linear head.
    \item We uncover LLMs' factual knowledge through \textbf{multiple probing strategies}, including yes/no queries and free-form generative questions.
    \item We estimate knowledge \textbf{encoded} in different LLMs' \textbf{outputs}, such as truth generation, fact generation, last layer activation, and logits.
    \item \textbf{Comprehensive} experiments on $4$ general-purpose LLMs, $3$ general knowledge datasets, and $7$ embedding models show that embeddings can effectively predict whether a fact is known by an LLM without querying it.
\end{enumerate}
\section{Related Work}\label{sec:relatedWork}

\paragraph{Factual knowledge benchmarks.} External resources, particularly, knowledge graphs, have been used to probe the factual knowledge learned by neural language models~\citep{youssef2023give}, by formulating the problem as a ``fill-in-the-blank'' cloze task for masked language models~\citep{petroni2019language,jiang2020canknow,kandpal2023large} 
and soon extended to large language models using cloze tasks, question-answering tasks, and open-ended generation tasks~\citep{luo2023systematic,sun2023head,kandpal2023large,min2023factscore,bai2024kgquiz}.
These benchmarks involve carefully selecting test facts from the vast sources using different techniques such as the popularity of the involved entities~\citep{sun2023head} and Thomson sampling~\citep{zheng2023kglens}. 
Thus, they only probe a subset of the factual knowledge, while PEEK enables us to uncover the factual knowledge without querying by learning proxy models. 

\noindent \paragraph{Non-factuality prediction and detection.} 
LLMs have been shown to contain multiple factual gaps, which leads them to confabulate/hallucinate incorrect information in an unpredictable manner~\citep{huang2023hallucination}.
To understand this phenomenon, existing works have investigated their intermediate representations and found that they can be used to predict the truthfulness of a fact~\citep{azaria2023internal, li2023inference,liu2023cognitive}, how knowledgeable it is about an entity~\citep{gottesman2024estimating}, and to predict whether its output will be non-factual or not~\citep{wang2024faclens}. 
In addition, other works have also explored fine-tuning additional models for the specific task of detecting hallucinations~\citep{arteaga2024hallucination,chen2023hallucination}. 
Hallucinations have also been associated with the model's epistemic uncertainty about a given fact~\citep{ahdritz2024distinguishing}. Existing works thus aim to estimate the uncertainty by checking inconsistencies in repeated samples~\citep{manakul2023selfcheckgpt}, teaching models to generate uncertainty~\citep{lin2022teaching}, using a bigger model's predictions as aleatoric uncertainty~\citep{ahdritz2024distinguishing}, and leveraging the logits for in-context examples~\citep{jesson2025estimating}.
Unlike these works, PEEK does not involve costly forward passes, a white-box access to the model, or extensive training of bulky modules. By adapting embedding models with a linear head, we can probe model's knowledge with respect to large databases without querying the bulky model. 

An extended discussion is provied in Appendix~\ref{app:relatedwork}. 
\section{Proxy Embeddings to Estimate Knowledge (PEEK)}\label{sec:method}
\begin{figure*}[t]
    \centering
    \includegraphics[width=0.9\linewidth]{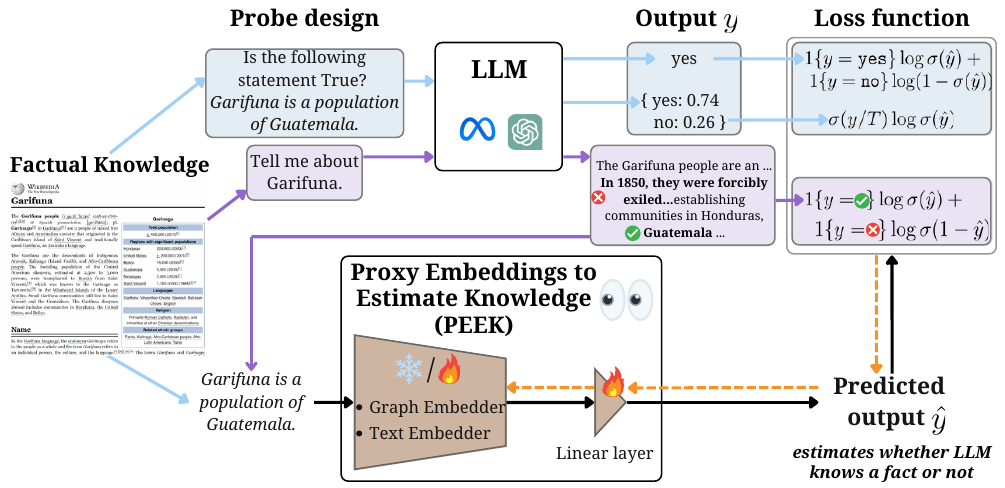}
    
    \caption{\textbf{Proxy Embeddings to Estimate Knowledge (PEEK):} In this framework, pre-trained embedding models are adapted to match the LLM knowledge for a training set of facts identified using different probing mechanisms. On a held-out set, we can then predict whether an LLM knows a fact or not by using the fact's embedding.}
    \label{fig:setup}
\end{figure*}

As shown in Figure~\ref{fig:setup}, we propose \textbf{PEEK} or proxy embeddings to estimate knowledge. More formally,

\begin{proposition}[\textbf{PEEK}]
    We are given a large language model $\CM$ and a knowledge base (a set of facts) $\CK$. Let $\CP_{\CM}: \CK \rightarrow \CO$ be a probing function that queries the LLM to determine if it knows a fact $f \in \CK$ and post-processes its output accordingly. Then, we propose that embeddings can be used as proxies for this function, \ie, $\hat{\CP}_{\theta} (f) = \Wbf^{\top} \CE (f) \approx \CP_{\CM}(f)$ for all facts $f \in \CK$, where $\CE$ is an embedding model that embeds facts $\in \CK$ to a low-dimensional space $\mathbb{R}^{d}$ and $\Wbf \in \mathbb{R}^{d}$. 
\end{proposition}
\subsection{Knowledge Probing Functions}
We consider four different ways of probing $\CP_{\CM}$ the LLM $\CM$ to find whether it knows a fact or not.

\noindent\paragraph{(1) Binary Generation.} 
The simplest way to probe the knowledge of a language model~\citep{clark2019boolq} is by querying it for the truthfulness of a fact. In particular, we prompt:
{\begin{quote}
\footnotesize
\texttt{You are only supposed to respond in yes/no.\\
Is the following statement <bool>?\\
STATEMENT: <fact>\\
ANSWER:
}
\end{quote}}
\noindent where \texttt{<bool>} is randomly sampled from $\{$\texttt{True, False}$\}$, and \texttt{<fact>} is a fact sampled from a knowledge database. 
For example, 
``Garifuna is a population of Guatemala''
is a valid \texttt{<fact>}. 
The generated response can be then post-processed into a binary label $y \in \CO = \{0, 1\}$, where $y=1$ corresponds to the LLM knowing the fact, \ie, if the output is `\texttt{yes}'/`\texttt{no}' when \texttt{<bool>} is True/False. In the case of a negative fact (a fact is not actually true), $y=1$ when the output is `\texttt{yes}'/`\texttt{no}' and \texttt{<bool>} is False/True respectively. 

\noindent\paragraph{(2) Binary Logits Generation.} We consider the same setting as above but extract the logits of the expected token from the LLM, \ie, `\texttt{yes}'/`\texttt{no}' when \texttt{<bool>} is True/False. Thus, we obtain the probed values $y \in \CO = \mathbb{R}$, representing a measure of the LLM's ``external confidence'' for the fact being true/false based on \texttt{<bool>}. 

\noindent\paragraph{(3) Binary Activation Prediction.} Hidden representations are known to possess knowledge about the truthfulness~\citep{azaria2023internal}. Thus, we probe this knowledge directly by prompting it with \emph{just} the \texttt{``<fact>''}  and training a linear head on top of a fixed layer of the last token of the LLM to predict $1$ if \texttt{<fact>} is true and $0$ if it is a negative fact. We thus train the linear layer for $10$ epochs on $80\%$ of the data. Then, we predict the logits of these binary predictions from the linear layer for the complete dataset, \ie, $y \in \CO = \mathbb{R}$. Thus, it measures the ``internal confidence'' of an LLM for a fact to be true/false. 

\noindent\paragraph{(4) Fact Generation.}
Finally, we also consider a more implicit way of probing the knowledge, where it is asked to generate a long-form biography or description of an entity/person. 
For example, ``Tell me about Garifuna''.
Here, we leverage an existing dataset, FactScore~\citep{min2023factscore} that extracts these passages from LLMs and then label each atomic facts from the passage based on whether it is supported, not supported, or irrelevant corresponding to a knowledge base. 
By considering the supported facts as $1$ and the unsupported facts as $0$, we get a dataset of atomic facts \texttt{<fact>} generated by LLMs that are either known by the LLM or not, labeled as
$y \in \CO = \{0, 1\}$. 

\subsection{Proxy Embeddings}
To estimate the LLM-probed knowledge, we use two different pre-trained embedding models that encode this knowledge in a self-supervised way.

\noindent\paragraph{(1) Sentence embedding models.}
Facts are sentences that combine entities with specific relations. 
Thus, one can find meaningful representations of them by leveraging pre-trained sentence embedding models~\citep{behnamghader2024llm2vec,muennighoff2022mteb}. These models are trained to represent ``similar'' sentences in a database close to each other. Two sentences can be deemed ``similar'' based on relevance for retrieval~\citep{thakur2021beir} and clustering tasks~\citep{rosenberg2007v}. While these models were traditionally trained from scratch on contrastive loss, newer models have adapted the progress in generative LLMs by leveraging their learned parameters for encoding sentences in a low-dimensional space. More details about training paradigms and architectural similarities are provided in Appendix~\ref{app:extended_discussion}.

\noindent\paragraph{(2) Graph neural networks.}
It has been shown that the reasoning abilities of the LLMs likely emerge due to a graphical combination of knowledge within texts~\citep{wang2024understanding}.
Facts sampled from a knowledge graph are structured as a triple consisting of a head entity, a relation, and a tail entity. 
Thus, we can exploit a Graph Neural Network (GNN) encoder to embed each fact of a knowledge graph in a low-dimensional space.
These are trained on a self-supervised loss to predict whether an edge exists for a head, relation, or tail triple. 
We employ a pre-trained foundational model for knowledge graphs, ULTRA~\citep{galkin2023ultra}, which encodes entities using the shortest path on the graph. 
This ensures that triples with an edge in the graph are separated in the embedding space from those that do not form an edge. 

\subsection{Proxy Tuning}
Since embeddings are not pre-trained to estimate probed knowledge from LLMs, they cannot be used directly. 
Thus, we adapt them by tuning their parameters to minimize a loss function that matches the LLM-probed knowledge with the embedding-based prediction for a small training dataset. 
Then, we consider two common approaches to efficiently tune the parameters~\citep{bommasani2021opportunities}:
\begin{itemize}[itemsep=0mm]
    \item \textbf{Low-rank tuning:} We update the parameters of the embedding models by adding a low-rank matrix~\citep{hu2021lora}, \ie, $\Wbf^T\CE + \alpha/r \Delta \Thetabf$, where $\Delta \Thetabf$ has rank at most $r$.
    \item \textbf{Linear-tuning:} One can also simply update the final linear layer $\Wbf$ while keeping the parameters of the embedding $\CE$ frozen. 
\end{itemize}

\textbf{\textit{Loss function}.} We tune these parameters to minimize the loss function such that $\CP_{\CM}(f) \approx \hat{\CP}_{\theta}(f)$ for all $f \in \CK$. Depending on the probing function $\CP$, we consider different losses. When $\CP_{\CM} \rightarrow \CO = \{0, 1\}$, \ie, for binary generation and fact generation probing, we use the binary cross-entropy loss: $\sum_{f} \CP_{\CM}(f)\log \sigma(\hat{\CP}_{\theta}(f)) + (1 - \CP_{\CM}(f))\log (1 - \sigma(\hat{\CP}_{\theta}(f)))$, where $\sigma$ is the sigmoid function. On the other hand, we use the knowledge distillation loss with temperature $T$: $\sum_{f} \sigma(\CP_{\CM}(f)/T) \log(\sigma(\hat{\CP}_{\theta}(f)/T)$,
when $\CP_{\CM} \rightarrow \CO = \mathbb{R}$, \ie, for binary logits generation and binary activation prediction.







\section{Experimental Setup}\label{sec:setup}

\paragraph{Datasets.} For binary prediction, logits generation, and activation prediction, we construct datasets of boolean knowledge questions by downsampling large knowledge graphs, specifically DBP100k~\citep{lehmann2015dbp100k} (697572 facts) and YAGO310~\citep{suchanek2007yago310} (1089040 facts), 
by considering 10\%, 1\%, and 0.1\% of the graph. Furthermore, we ensure that the proportion of facts with a particular relation type is preserved in the downsampled set to avoid sampling from a particular relation type only.
Since knowledge graphs encode semantic relations differently from natural language, direct decoding into natural language is non-trival~\citep{gardent2017creating,gardent2017webnlg}. Thus, we employ a heuristic approach by manually annotating a query prompt for each relation type $r$ using placeholders for the head ($h$) and tail ($t$) entities. This is an efficient strategy to generate textual prompts since common knowledge graphs often contain only hundreds of relation types compared to millions of factual relationships. Some examples of templates are provided in Table~\ref{tab:templates} (Appendix ~\ref{app:relation}). 

For fact generation task, we use the ChatGPT-generated labels of FactScore for different labels~\citep{min2023factscore} due to their high alignment with humans. 
We split all datasets into training (0.8), validation (0.1), and test (0.1) sets with respective splits and report the results on the held-out test set. 
Note that since we randomly split the dataset, the test splits can include new nodes testing for the inductiveness of the entity representation. For more details, refer Appendix~\ref{app:details}.


\noindent\paragraph{Embedding models.} For LLM knowledge estimation on knowledge graphs, we consider a pre-trained graph neural network ULTRA~\citep{galkin2023ultra} while we consider 6 state-of-the-art other sentence embedders for generic LLM knowledge estimation: 
MPNET~\cite{song2020mpnet},  
NVE2~\cite{leenv}, 
Linq~\cite{LinqAIResearch2024}, 
GTE~\cite{li2023towards}, 
GIST~\cite{solatorio2024gistembed}, 
and MXBAI~\cite{li2023angle}. 
These are chosen for their high performance due to their high retrieval performance on knowledge-related tasks on MTEB~\cite{muennighoff2023mteb} benchmark. 

\noindent\paragraph{Large language models.} For knowledge graphs, we estimate the knowledge of 1 open-source model: Llama-3.1-8B~\cite{dubey2024llama} 
as well as 2 proprietary models: GPT-4o and GPT-4o-mini~\cite{openai2023gpt4}. 
We also consider the proprietary models to reason about our prompt with a standard chain-of-thought prompt (COT)~\citep{kojima2022large}: \texttt{Think step-by-step} and acquire the final result using structured outputs API~\footnote{\url{https://platform.openai.com/docs/guides/structured-outputs\#chain-of-thought}}. For FactScore, we use the LLMs used in their paper, \ie, we consider the labeled facts generated by Alpaca~\citep{taori2023stanford}, StableLM-alpha~\cite{bellagente2024stable}, 
ChatGPT~\citep{openai2022chatgpt}, Vicuna~\citep{chiang2023vicuna}, and InstructGPT~\citep{ouyang2022training}. 

\noindent\paragraph{Metrics.} When the LLM probing outputs are binary (\ie, for binary generation and fact generation), we consider accuracy (ACC) and area under ROC (AUC) of the trained estimation model. On the other hand, when the outputs are real-valued (\ie, for binary logits generation and binary activation prediction), we use mean absolute error (MAE) between the predicted and actual values. 

\noindent \paragraph{Baselines.} We consider three different baselines to show the efficacy of using proxy embeddings. In particular, we employ the Majority classifier (\ie, always predicting the majority class) and a Random classifier (\ie, randomly predicting the class) for the binary generation task. In addition to this, we follow ~\citep{azaria2023internal} and employ LlamaHid, that uses the hidden representations of the Llama-3.1-8B at layer $10, 20, 30$ and compare the best performance.

\noindent \paragraph{Hyperparameters.} For linear tuning, we find the best embedding model using performance on the validation set for the learning rate $\in [10^{-3}, 10^{-2}]$ and the number of epochs $\in [20, 40]$. On the other hand, we consider number of epochs $\in [5, 10]$ for LoRA tuning with $\alpha = 8$ and $r = 8$. 

\section{Results}\label{sec:experiments}
\begin{table*}[t]
    \centering
    \resizebox{1.0\textwidth}{!}{
    \begin{tabular}{cccHHccccccccHc}
    \toprule
     \multicolumn{15}{c}{DBP-100k} \\
     \hline
     & \multicolumn{2}{c}{Llama-3.1-8B} & & & \multicolumn{2}{c}{GPT-4o-mini} & \multicolumn{2}{c}{GPT-4o} & \multicolumn{2}{c}{GPT-4o-mini COT} & \multicolumn{2}{c}{GPT-4o COT} & \multirow{2}{*}{Average rank for AUC+ACC} & \multirow{2}{*}{\shortstack{Average rank for \\ AUC+ACC$^*$}} \\
     & AUC $\uparrow$ & ACC $\uparrow$ & AUC $\uparrow$ & ACC $\uparrow$ & AUC $\uparrow$ & ACC $\uparrow$ & AUC $\uparrow$ & ACC $\uparrow$ & AUC $\uparrow$ & ACC $\uparrow$ & AUC $\uparrow$ & ACC $\uparrow$ &  &  \\
    \midrule
    Majority & - & 67.23 & - & 52.40 & - & 66.67 & - & 69.49 & - & 76.27 & - & 83.05 & 9.00 & 9.00 \\
    Random & - & 42.94 & - & 51.29 & - & 46.33 & - & 50.85 & - & 44.63 & - & 49.15 & 10.00 & 10.00 \\
    Llama-Hid & 60.70 & 73.45 & \textbf{67.01} & \textbf{61.62} & 61.67 & 66.10 & 58.59 & 73.45 & 60.58 & 67.23 & 65.86 & 80.79 & 6.33 & 7.40 \\
    \hline
    MPNET & 66.66 & 80.94 & 51.14 & 51.47 & 73.28 & 88.53 & 68.06 & 87.57 & 73.01 & 87.79 & 69.50 & 90.00 & 4.50 & 5.00 \\
    NVE2 & \textbf{80.80} & \textbf{81.71} & 51.07 & 50.63 & \textbf{86.47} & 90.22 & 80.83 & 87.50 & \textbf{88.82} & 90.16 & 86.89 & 91.52 & 1.83 & \textbf{1.40} \\
    GIST & 51.11 & 73.45 & 51.23 & 50.81 & 71.69 & 90.42 & 70.07 & 88.65 & 71.85 & 87.28 & 70.18 & 90.81 & 5.33 & 5.80 \\
    Linq & 64.53 & 73.45 & 47.51 & 50.48 & 85.12 & \textbf{91.09} & \textbf{82.01} & \textbf{88.11} & 86.88 & \textbf{90.53} & \textbf{88.72} & \textbf{91.62} & 3.00 & 2.00 \\
    GTE & 62.03 & 73.45 & 49.79 & 50.76 & 81.73 & 90.42 & 76.59 & 87.57 & 85.11 & 88.77 & 85.56 & 91.49 & 3.67 & 3.40 \\
    MXBAI & 51.93 & 73.45 & 49.45 & 50.88 & 73.29 & 90.42 & 70.53 & 88.65 & 73.79 & 87.28 & 70.93 & 90.95 & 4.83 & 4.60 \\
    ULTRA & 58.58 & 80.80 & 49.45 & 49.35 & 66.32 & 88.73 & 60.08 & 71.62 & 64.82 & 87.35 & 64.78 & 90.00 & 6.50 & 6.40 \\
    \midrule
    \midrule
    \multicolumn{15}{c}{YAGO310} \\
    \hline
    & \multicolumn{2}{c}{Llama-3.1-8B} & &  & \multicolumn{2}{c}{GPT-4o-mini} & \multicolumn{2}{c}{GPT-4o} & \multicolumn{2}{c}{GPT-4o-mini COT} & \multicolumn{2}{c}{GPT-4o COT} & \multirow{2}{*}{Average rank for AUC+ACC} & \multirow{2}{*}{\shortstack{Average rank for \\ AUC+ACC$^*$}} \\
     & AUC $\uparrow$ & ACC $\uparrow$ & AUC $\uparrow$ & ACC $\uparrow$ & AUC $\uparrow$ & ACC $\uparrow$ & AUC $\uparrow$ & ACC $\uparrow$ & AUC $\uparrow$ & ACC $\uparrow$ & AUC $\uparrow$ & ACC $\uparrow$ &  &  \\
    \midrule
    Majority & - & 68.00 & - & 50.72 & - & 52.89 & - & 51.24 & - & 54.55 & - & 60.33 & 9.00 & 9.00 \\
    Random & - & 47.56 & - & 50.45 & - & 51.24 & - & 47.93 & - & 52.07 & - & 52.89 & 10.00 & 10.00 \\
    Llama-Hid & 59.33 & 67.55 & \textbf{58.77} & \textbf{52.89} & 79.67 & \textbf{70.25} & \textbf{66.51} & 59.50 & 79.83 & 72.72 & 70.63 & 63.64 & 2.50 & 2.80 \\
    \hline
    MPNET & 59.08 & 64.44 & 48.50 & 51.14 & 77.50 & 67.77 & 60.80 & 57.02 & 75.77 & 67.77 & 64.73 & 61.98 & 5.50 & 5.60 \\
    NVE2 & 60.56 & 65.22 & 50.55 & 51.06 & \textbf{82.97} & 69.42 & 64.65 & 57.85 & \textbf{81.81} & \textbf{73.55} & 65.92 & 60.06 & 3.50 & 3.60 \\
    GIST & 65.80 & 62.40 & 50.70 & 51.23 & 73.08 & 63.64 & 61.56 & 54.55 & 73.98 & 66.12 & 68.27 & 61.16 & 5.67 & 6.40 \\
    Linq & \textbf{69.16} & 65.18 & 47.61 & 47.51 & 82.75 & 68.60 & 64.05 & \textbf{60.33} & 78.43 & 71.07 & \textbf{78.13} & \textbf{70.25} & 2.83 & \textbf{1.80} \\
    GTE & 59.97 & \textbf{71.11} & 48.94 & 49.79 & 80.47 & 64.46 & 65.04 & 55.37 & 77.42 & 67.77 & 77.13 & \textbf{70.25} & 4.17 & 3.60 \\
    MXBAI & 59.33 & 70.22 & 50.65 & 49.45 & 74.07 & 64.46 & 62.38 & 55.37 & 75.30 & 66.94 & 69.24 & 62.81 & 5.17 & 5.40 \\
    ULTRA & 56.17 & 64.33 & 49.75 & 49.45 & 75.44 & 65.29 & 62.79 & 58.68 & 72.01 & 66.12 & 66.97 & 58.68 & 6.67 & 6.80 \\
    \bottomrule
    \end{tabular}}
    \caption{\textbf{Binary Generation:} Accuracy (ACC) and AUC for predicting LLM-generated binary truth for $0.1\%$ true facts sampled from knowledge graphs. Standard deviation $\le 1$ for accuracy and $\le 5$ for AUC. 
    }
    \label{tab:binary_gen}
\end{table*}


\subsection{Binary Generation}

\subsubsection{Base performance}\label{sec:base}
First, we establish the hardness of our task of estimating the LLM-acquired knowledge by validating their accuracy for the binary generation task when compared with the true factual knowledge of the two datasets. Table~\ref{tab:base_accuracy} (App.~\ref{app:results}) shows that around 40-70 \% of the true facts are accurately predicted by the model in each dataset for different sampling percentages assuming 0 negative samples. This also show that the binary labels of knowledge are \textbf{\emph{verified to be balanced}}, making the accuracy and AUC metrics suitable for this evaluation and using a majority or random classifier a suboptimal choice.

Then, we estimate the LLM's knowledge using the binary generation probing strategy on large knowledge graphs (\ie, DBP100k and YAGO310) by downsampling the graphs to $0.1\%$ of the total size and only probing the positive facts (\ie, triples with an edge). 
Table~\ref{tab:binary_gen} shows the effect of linear tuning of different embedding models to match the binary generation probes from different LLMs on the test set. 
In almost all cases, we find that sentence embedding models are more effective at estimating the knowledge acquired by different LLMs than all the baselines including the strong and arguably more expensive Llama-Hid, achieving up to $91\%$ accuracy and up to $88\%$ AUC scores for GPT-4o-mini. This highlights the effectiveness of PEEK in efficiently and accurately identifying factual gaps in LLMs compared to existing methods. 
We find that, in most cases, embedding models are very effective in estimating LLMs' knowledge in these datasets. Further analysis in Appendix~\ref{app:results} shows that a notable exception is Llama3.3-70B, for which all embedding models fail to estimate its knowledge.  has almost random accuracy in finding the correct knowledge over these datasets, indicating that estimating it is effectively as difficult as predicting the output of a random number generator. 

Nvidia's NVE2 is the overall winner in DBP100k while Linq performs the best and second-best in YAGO310 and DBP100k for different LLMs. 
This shows that Linq is best at estimating the knowledge acquired by arbitrary LLMs. 
However, a deeper look at the results shows that the ranks are mainly due to high performance on GPT-4o-mini and the results on GPT-4o in YAGO310 and Llama models are just above random performance. This shows that there is significant room for improvement for PEEK. Another interesting observation is that pre-trained representations of a graph neural network, ULTRA, does not align with what LLMs know to be true/false, as ULTRA performs the worst among all considered embedding models. 
This shows that 
LLMs do not induce facts in the same way as a path-based graph neural network does on a knowledge graph, \ie, using shortest paths in a Wikipedia-based graph.

\begin{figure*}[t]
    \centering
    \includegraphics[scale=0.7]{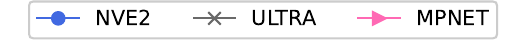} \\
    \hspace*{\fill}
    \subfloat[DBP100k 4o]{\includegraphics[width=0.22\textwidth]{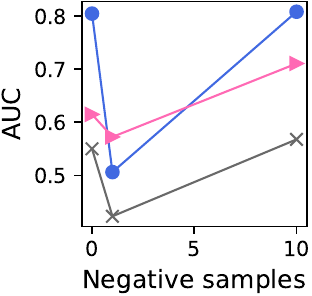}} \hfill
    \subfloat[DBP100k 4o-mini]{\includegraphics[width=0.22\textwidth]{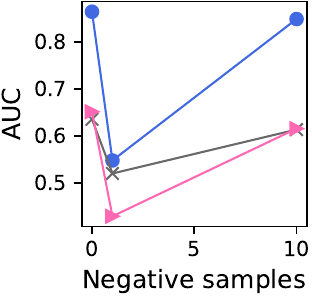}} \hfill
    \subfloat[YAGO310 4o]{\includegraphics[width=0.22\textwidth]{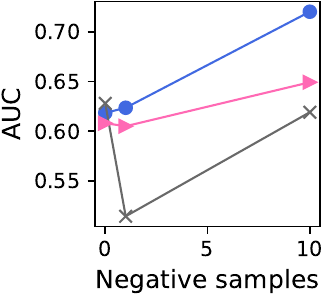}} \hfill
    \subfloat[YAGO310 4o-mini]{\includegraphics[width=0.22\textwidth]{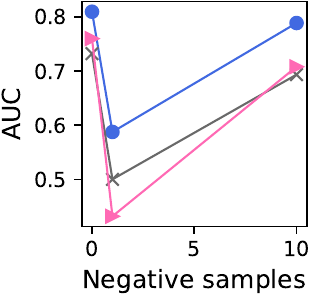}} 
    \hspace*{\fill}
    \caption{Effect of changing the number of negative samples in GPT models for knowledge graphs.}
    \label{fig:negative_samples}
\end{figure*}

\begin{figure*}[t]
    \centering
    \includegraphics[scale=0.7]{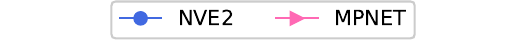} \\
    \hspace*{\fill}
    \subfloat[DBP100k]{\includegraphics[width=0.2\textwidth]{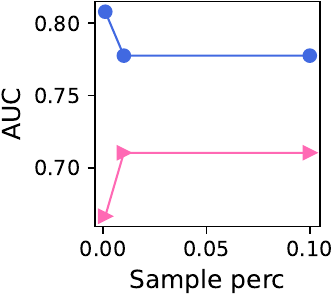}} \hfill
    \subfloat[DBP100k]{\includegraphics[width=0.2\textwidth]{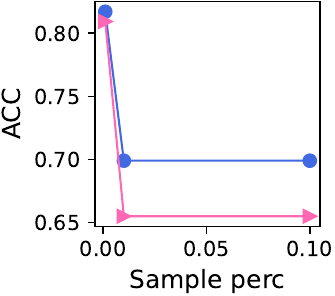}} \hfill
    \subfloat[YAGO310]{\includegraphics[width=0.2\textwidth]{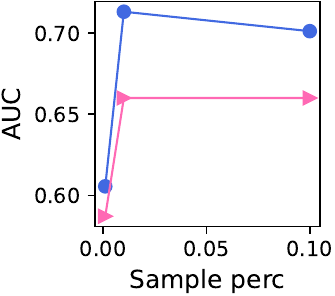}} \hfill
    \subfloat[YAGO310]{\includegraphics[width=0.2\textwidth]{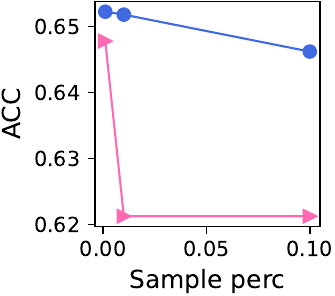}} 
    \hspace*{\fill}
    
    \caption{Effect of changing the number of negative samples in Llama3.1-8B for knowledge graphs. 
    }
    \label{fig:sample_perc}
\end{figure*}

\begin{figure}[t]
    \centering
    \hspace*{\fill}
    \subfloat[GPT-4o]{\includegraphics[width=0.22\textwidth]{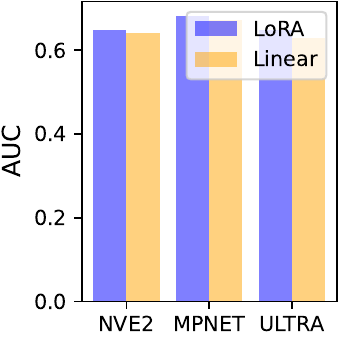}} \hfill
    \subfloat[GPT-4o-mini]{\includegraphics[width=0.22\textwidth]{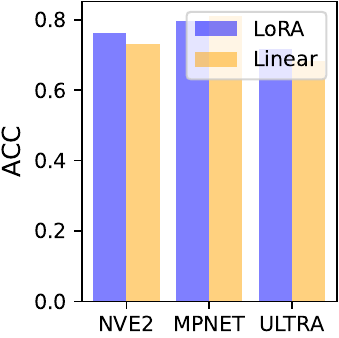}} \hfill
    \hspace*{\fill}
    \caption{\textbf{LoRA v/s Linear tuning} on $1\%$ DBP100k}
    \label{fig:tuning}
\end{figure}

\paragraph{Negative samples}
Next, we find how negative facts impact the estimation performance of the LLM knowledge in these knowledge graphs. In particular, for every positive fact/edge, $(h, r, t)$, we randomly sample a negative edge $(h, r, t')$ such that no edge exists corresponding to the triple. Figure~\ref{fig:negative_samples} shows how the AUC and accuracy changes of 3 representative embedding models: NVE2, ULTRA, and MPNET, as we change the number of negative samples/edges from $[0, 1, 10]$. We observe that AUC scores typically increase with an increase in the number of negative samples. Assuming LLMs know the ground truth, there is a likelihood of more negative facts and thus, the increase could be due to a natural class imbalance, and thus, the models tend to favor the majority class of ``false''. Furthermore, knowledge graphs tend to have missing knowledge, that is true but not modeled in the graph. To avoid these cases, we focus on $0$ negative samples.

\paragraph{Sampling percentage}
We also verify the stability of our results by varying the sampling percentage with a Llama3.1-8B model. Figure~\ref{fig:sample_perc} shows how AUC and accuracy vary in two representative embedding models (other models show similar results) for different sampling percentages. We can observe that the results are mostly preserved with a maximum change of $5\%$ in the AUC and $10\%$ in the ACC as we increase the sampling percentage. Thus, we consider a fixed sampling percentage of $0.1\%$ for our global results since it is more efficient and scalable to proprietary models, reducing API and computational costs.

\paragraph{Fine-tuning}

We also analyze if LoRA tuning improves the performance of embedding models as compared to linear tuning. Figure~\ref{fig:tuning} compares the AUC and accuracy of the three embedding models: NVE2, MPNET, and ULTRA for LoRA and linear tuning methods to estimate the knowledge of GPT-4o and GPT-4o-mini on $1\%$ of DBP100k dataset. One can observe that the LoRA tuning achieves identical performance as linear tuning, showing that LoRA tuning may not be needed for this task. This further strengthens the claim that embedding models implicitly encode knowledge in a way that can easily adapt to match an LLM's acquired knowledge space using just a linear layer.

\begin{table*}[t]
    \centering
    \resizebox{0.78\textwidth}{!}{
    \begin{tabular}{c c c c c c c c c}
        \toprule
        LLM & Metric & 
        MPNET & NVE2 & GIST & Linq & GTE & MXBAI \\
        \midrule
        \multirow{2}{*}{Alpaca} & AUC $\uparrow$ & 84.41 & \textbf{90.14} & 86.01 & 89.11 & 89.93 & 86.07 \\
         & ACC $\uparrow$ & 76.17 & 82.38 & 78.24 & \textbf{83.42} & 82.90 & 78.76 \\
        \midrule
        \multirow{2}{*}{StableLM} & AUC $\uparrow$ & 71.17 & 86.06 & 78.93 & 85.90 & \textbf{86.83} & 79.92 \\
         & ACC $\uparrow$ & 83.72 & 84.50 & 83.72 & 84.88 & \textbf{86.05} & 83.72 \\
        \midrule
        \multirow{2}{*}{ChatGPT} & AUC $\uparrow$ & 67.88 & \textbf{79.73} & 68.04 & 78.90 & 77.56 & 70.15  \\
        & ACC $\uparrow$ & 76.51 & \textbf{80.63} & 74.92 & 79.37 & 78.10 & 75.24 \\
        \midrule
        \multirow{2}{*}{Vicuna} & AUC $\uparrow$ & 81.54 & 86.93 & 82.99 & \textbf{88.19} & 87.26 & 83.16 \\
        & ACC $\uparrow$ & 74.18 & 79.08 & 76.09 & 79.35 & \textbf{79.89} & 76.63 \\
        \midrule
        \multirow{2}{*}{InstructGPT} & AUC $\uparrow$ & 80.79 & \textbf{84.80} & 80.94 & 84.64 & 84.58 & 80.74 \\
        & ACC $\uparrow$ & 76.56 & 76.92 & 72.89 & \textbf{77.66} & 76.19 & 73.63 \\
        \midrule
        \multicolumn{2}{c}{Overall rank for $\tfrac{1}{2}$(AUC+ACC)} & 6 & 2 & 5 & 1 & 3 & 4 \\
        \bottomrule
    \end{tabular}}
    \caption{\textbf{Fact Generation:} Accuracy (ACC) and AUC for predicting (in)correct facts generated by LLMs.}
    \label{tab:factscore}
\end{table*}

\begin{table}[t]
    \centering
    \resizebox{0.5\linewidth}{!}{
    \begin{tabular}{c c c }
        \toprule
        Embedding & DBP-100k & YAGO-310 \\
        \midrule
        MPNET &  4.30 & 1.78 \\
        NVE2 & 2.98 & \textbf{1.64} \\
        GIST & 2.60 & 1.81 \\
        Linq & \textbf{2.36} & 1.67 \\
        GTE & 2.88 & 1.69 \\
        MXBAI & 2.43 & 1.84 \\
        \bottomrule
    \end{tabular}
    }
    \caption{\textbf{Binary logits generation:} Mean absolute error in predicting the Llama3.1-8B logits of the correct token for $1\%$ data. Lesser is better.}
    \label{tab:binary_logits}
\end{table}
\begin{table}[t]
    \centering
    \resizebox{0.5\linewidth}{!}{
    \begin{tabular}{c c c }
        \toprule
        Embedding & DBP-100k & YAGO-310 \\
        \midrule
        MPNET &  0.5714 & 0.5722 \\
        NVE2 &  \textbf{0.4746} & 0.5006 \\
        GIST &  0.5724 & 0.5392 \\
        Linq &  0.4845 & 0.4895 \\
        GTE &  0.4959 & \textbf{0.4847} \\
        MXBAI &  0.5625 & 0.5353 \\
        \bottomrule
    \end{tabular}
    }
    \caption{\textbf{Binary activation prediction:} Mean absolute error in predicting the adapted logits from the activation layer of Llama3.1-8B on $0.1\%$ data. Lesser is better.}
    \label{tab:binary_activation}
\end{table}

\subsection{Binary Logits Generation}

Here, we directly adapt the embedding models to predict the logits of the correct token for a prompt (\ie, ``yes'' token when \texttt{<bool>} is True and ``no'' token when \texttt{<bool>} is False) using the knowledge distillation loss. Table~\ref{tab:binary_logits} reports the mean absolute error between the predicted and actual logits of Llama3.1-8B for different embeddings, as we search over temperatures $T \in [1, 5, 10]$. We can observe the high performance of Linq model in this case as well with NVE2 and GTE models coming close. However, the error is still quite high (all $> 1$), leaving room for enough improvement in estimating the LLM's logit-level knowledge. 

\subsection{Binary Activation Prediction}

Since knowledge can be encoded in the representations and not get translated to the final logits~\citep{azaria2023internal}, we also train an additional linear layer on the hidden activations of Llama3.1-8B to separate true and negative facts in a knowledge graph. 
We use 1 and 10 negative samples for training and report the best results in Table~\ref{tab:binary_activation} compares the estimation using the last hidden layer over temperatures $T \in [1, 5, 10]$. We observe similar trends as above and note that error values are smaller due to more normalized logits tuned for the binary classification using the adapted linear layer. NVE2 and GTE perform the best for the two datasets. We further extend this analysis by using an intermediate 15th layer in Table~\ref{tab:binary_activation001} (Appendix~\ref{app:extended_discussion}) and find the trends for $0.1\%$ and $1\%$ data to be similar. 

\subsection{Fact Generation}
Table~\ref{tab:factscore} shows the performance of embedding models to estimate the knowledge of different LLMs as found through open-ended fact generation. We report the accuracy and AUC scores on the test set by learning to label the correct facts generated by the LLM as $1$ and the incorrect facts as $0$. Since these facts can be formed from arbitrary English text, we omit the graph neural network, ULTRA from this comparison. We find that the embedding models can estimate this knowledge space as well with upto $90\%$ AUC and $86\%$ accuracy. Results further show that Linq outperforms all the other embedding models with NVE2 coming second. This complements our findings on the binary generation task and establishes that these embeddings are quite effective in adapting to the knowledge space of different LLMs with a linear layer across different knowledge probing strategies.

\section{Conclusion}
\label{sec:conclusion}
We presented PEEK, a novel framework for estimating knowledge encoded in large language models using proxy embedding models. 
Our method offers a scalable and cost-effective way to assess what an LLM knows, which is essential for identifying and addressing factual gaps before deployment.
Embedding models should thus be used in the pre-deployment stage of any LLM to pre-emptively find and fix factual gaps in the LLMs. 
Effective estimation of underlying knowledge using embeddings enables us to efficiently utilize the context in retrieval-augmented generation efficient context utilization by disregarding already-known facts.
Another interesting future direction lies in analyzing how the proxy embeddings can dynamically adapt as the model continues to learn from limited examples. 

\section*{Limitations}\label{sec:limitations}

\textbf{Confounding of other factors.} A limitation of our work is that our results can be confounded by a lot of hidden factors in the learning process and data of the two models. While these findings show significant promise, it is likely that GPT-generated knowledge or architecture itself~\citep{behnamghader2024llm2vec} is leaked to train embedding models. Since the main contribution of our work is in identifying the potential of embedding models to estimate LLM knowledge, these factors do not undermine our results in efficient factual probing. We thus leave it for future works to analyze the mechanisms of why certain embeddings estimate knowledge better than others and provide a starting point in Section~\ref{app:extended_discussion}.

\noindent \textbf{Other probing tasks.} Another limitation is that we do not consider more difficult probing tasks such as fill-in-the-blank or multiple choice question-answering since they can lead to arbitrary generation of facts which are harder to estimate through encoder-only models. 

\noindent \textbf{Wikipedia.} While our benchmark focuses on Wikipedia-derived knowledge (e.g. DBP100k), we emphasize that our method is not inherently restricted to factual or closed-domain settings. We also note that harder domain-specific knowledge would potentially require training domain-specific embeddings from scratch since existing embedding models may not necessarily generalize to these. Our focus here is to show a standardized probing and estimation pipeline and we believe this would generalize well to these settings as well given such specialized embeddings exist. While Wikipedia is a widely accepted source of knowledge~\citep{thorne2018fever,yang2018hotpotqa}, more specialized sources can be explored in the future. 





\bibliographystyle{unsrt}
\bibliography{citations}

\clearpage

\appendix
\appendix
\section*{Appendix}





\section{Extended Related Work}~\label{app:relatedwork}

\noindent\textbf{Interpretable LLM representations.} 
Prior work has explored the problem of understanding the internal mechanism of these extremely large models by learning sparse autoencoders~\citep{ng2011sparse,lieberum2024gemma}, prompting-based explanations~\citep{zhao2024explainability}, and understanding induction on simpler tasks~\citep{nogueira2021investigating,quirke2023understanding,wang2024understanding}. In the same vein, the interpretability of factual knowledge induction in LLMs can also be improved by leveraging smaller proxy models with well-defined inductive biases to estimate LLM knowledge.

\noindent \textbf{Applications.} In \emph{retrieval-augmented generation} (RAG)
~\citep{khandelwal2019generalization,guu2020retrieval,lewis2020retrieval}, where managing long contexts is a key challenge~\cite{yoon2024compact, cheng2024lift, xu2023recomp}, 
PEEK can filter out already-known facts to the LLM, reducing context length and improving retrieval efficiency. 
For \emph{hallucination detection}~\citep{li2023halueval}, existing works explored LLMs' self-awareness of their own hallucinations~\citep{liu2023cognitive,duan2024llms} and the use of fine-tuned auxiliary models~\citep{arteaga2024hallucination} for uncertainty estimation~\citep{vashurin2024benchmarking}. 
Our approach complements prior work on hallucination detection by providing a scalable proxy for evaluating LLM self-knowledge. 
In \emph{model editing and updating}, where factual knowledge is localized and modified within the model~\citep{wallat2021bertnesia,roberts2020howmuch,meng2022locating,farquhar2023challenges, meng2022locating,mitchell2021fast}, PEEK embeddings can help identify misaligned facts, which can then be corrected via neuron-level editing~\citep{meng2022locating,mitchell2021fast} or knowledge distillation~\citep{padmanabhan2024propagating}. 
Future works can also focus on directly aligning the LLMs to the knowledge space of facts by using embedding alignment techniques or reinforcement learning~\citep{lee2023rlaif}. Finally, we flag that training embedding models specifically to estimate LLM knowledge can be a double-edged sword as it may lead to forgetting more useful true knowledge, resulting in bad performance on retrieval tasks. We leave it for future works to study this trade-off in more detail and how both fields can grow hand-in-hand. 

\section{Additional Experimental Details}

\subsection{Relation Annotation}\label{app:relation}
We annotate each relation type to form a question template that encodes the semantic information of this relation. Since the number of relations are limited (at most 470), we can do this efficiently and effectively. Table~\ref{tab:templates} show some question templates used in this work while we provide the complete list in the supplementary material.

\subsection{Dataset Details}\label{app:details}

We also note that the data splits already include new nodes that were not seen during training. Table~\ref{tab:data_stats} shows that our splits contain novel entities in the test set that were not seen during training. 

\subsection{Implementation details}\label{app:implementation}
All the experiments were conducted on Python 3.8.12 on a Ubuntu 18.04 PC with an Intel(R) Xeon(R) CPU E5-2698 v4 @ 2.20GHz processor, 512 GB RAM, and Tesla A100 64 GB GPUs.

\section{Additional Experimental Results}\label{app:results}

\paragraph{Base accuracy.} Table~\ref{tab:base_accuracy} shows the base accuracy of different LLMs on the binary generation task for the two knowledge graphs at different splits. Here, we also include Llama-3.3-70B~\cite{dubey2024llama} for comprehensiveness. As discussed in the main paper, we include the spurious results of Binary generation on Llama-3.3-70B in Table~\ref{tab:binary_gen_llama70b}.

\begin{figure*}[t]
    \centering
    \hspace*{\fill}
    \subfloat[GPT-4o]{\includegraphics[width=0.24\textwidth]{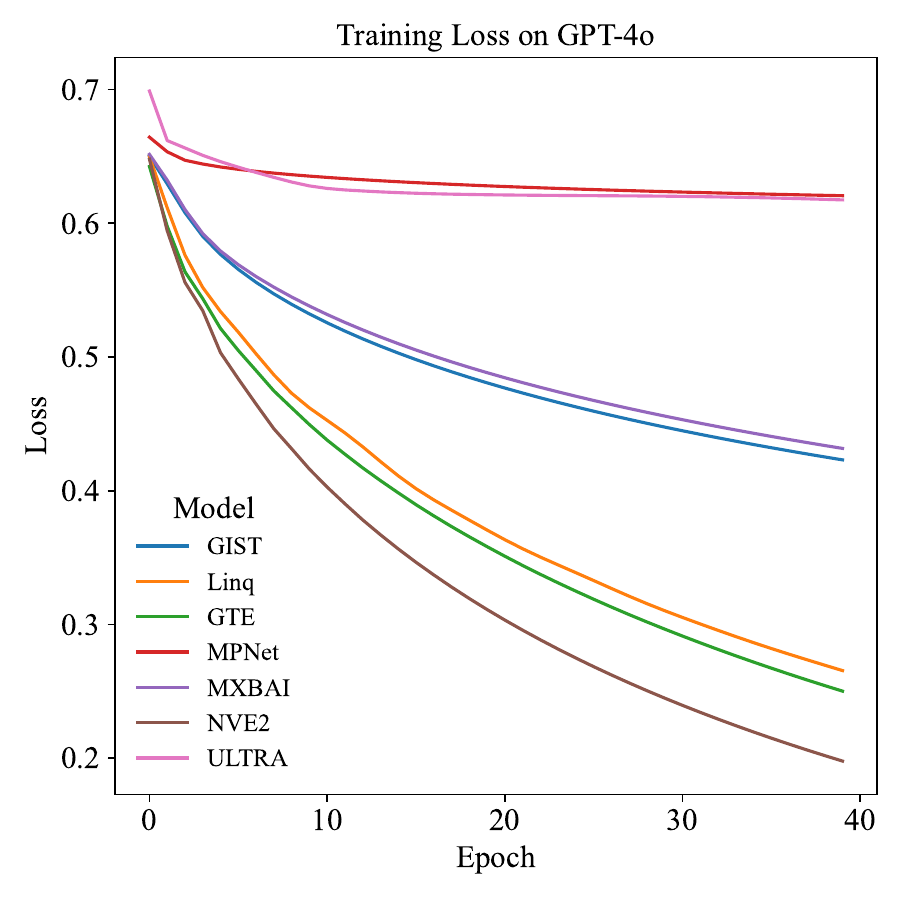}}
    \hfill
    \subfloat[GPT-4o COT]{\includegraphics[width=0.24\textwidth]{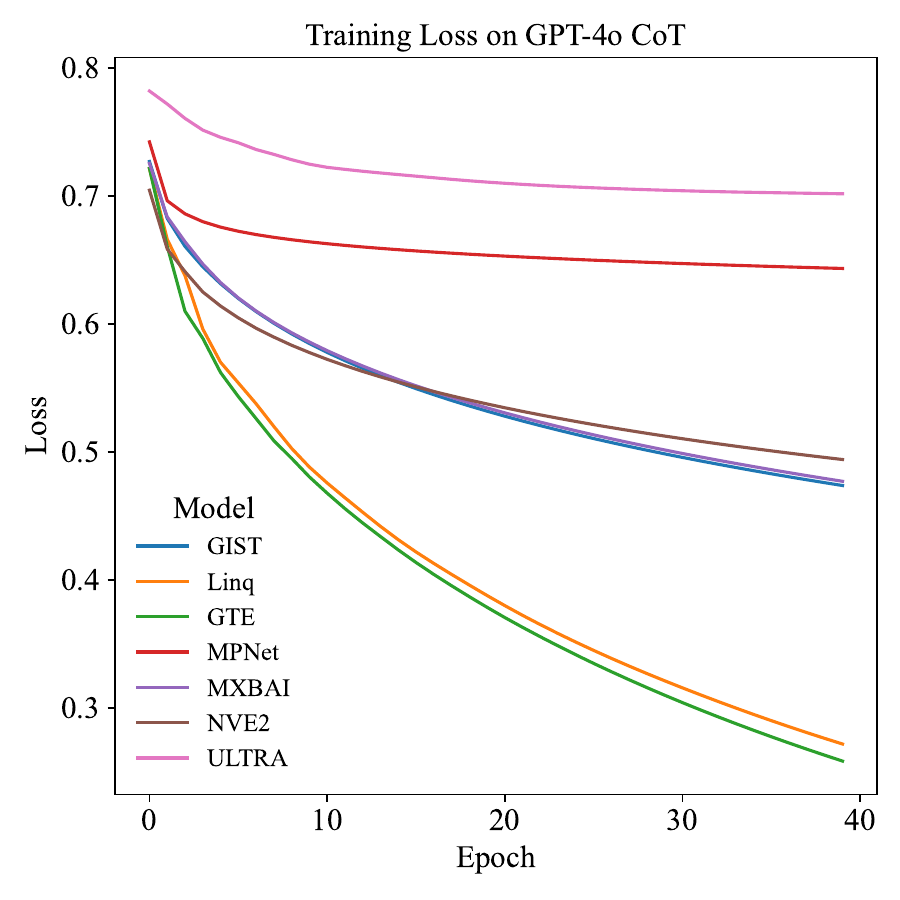}}
    \hfill
    \subfloat[GPT-4o-mini]{\includegraphics[width=0.24\textwidth]{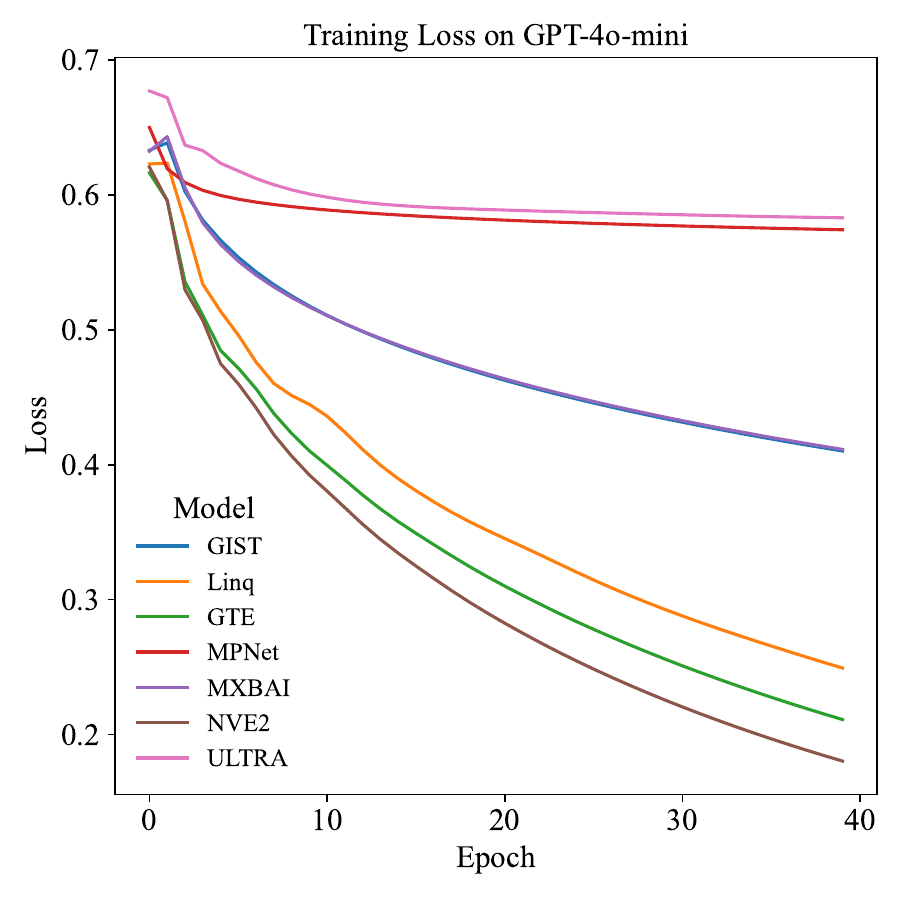}}
    \hfill
    \subfloat[GPT-4o-mini COT]{\includegraphics[width=0.24\textwidth]{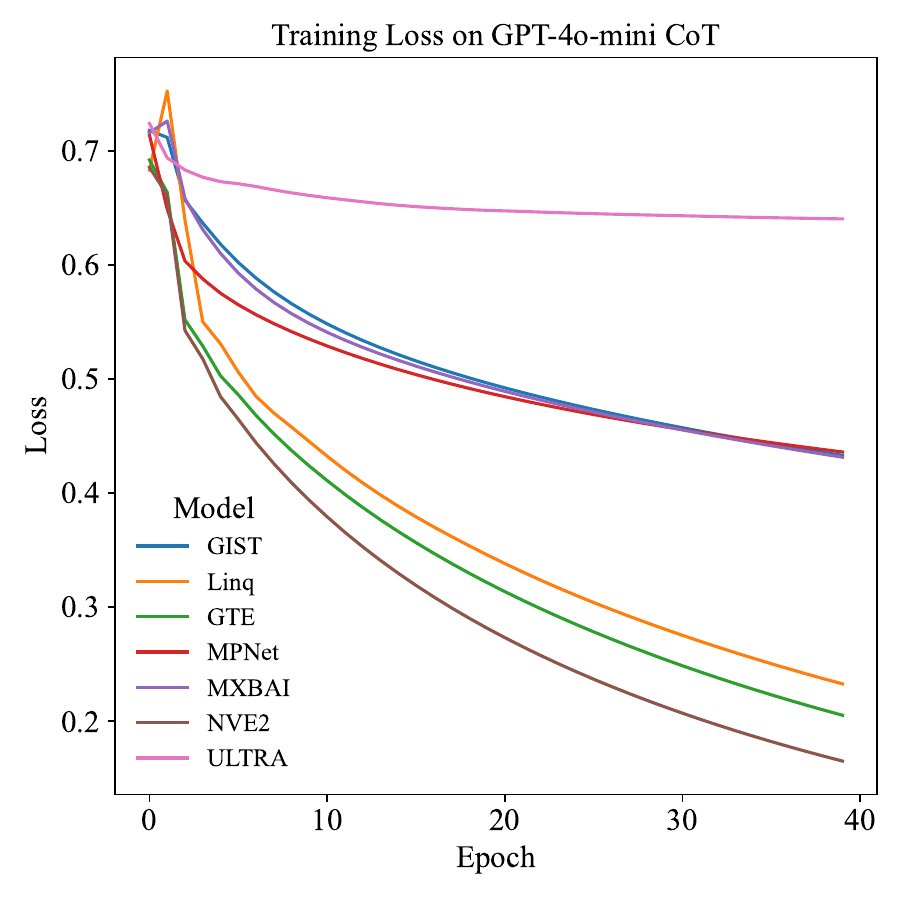}}
    \hspace*{\fill}
    \caption{\textbf{Training loss curves of various embedding models to adapt with the LLM knowledge probed using binary generation on $0.1\%$ DBP100k.}}
    \label{fig:loss_curves}
\end{figure*}

\paragraph{Training.}
Figure~\ref{fig:loss_curves} shows a set of representative loss curves for linear tuning of embeddings for binary generation in $0.1\%$ sampling of DBP100k. Similar curves are observed for other datasets and tasks.

\paragraph{Binary Activation Prediction.}
Table~\ref{tab:binary_activation001} shows the results on using the 15th layer of Llama-3.1-8B to probe for the knowledge. These results further complement our earlier findings of the effectiveness of the Linq embedding.

\paragraph{Examples.} Table~\ref{tab:examples} lists some examples of the correct and incorrect estimation by the best performing Linq embedding for GPT-4o in DBP100k. However, it is hard to make generalizable claims from the list.

\begin{table*}[tb]
    \centering
    \resizebox{0.9\textwidth}{!}{%
    \begin{tabular}{c c l}
        \toprule
        Dataset & Relation Type & Prompt Template \\
        \midrule
        \multirow{15}{*}{\shortstack{DBP100k \\ \\ (470 total \\ relation types)}} & associatedBand & \{t\} is an associated band of \{h\}. \\
        & instrument & \{t\} is the instrument of \{h\}. \\
        & ideology & \{t\} is the ideology of \{h\}. \\
        & country & \{h\} is in the country of \{t\}. \\
        & language & \{t\} is a language of \{h\}.\\
        & occupation & \{t\} has been the occupation of \{h\}.\\
        & education & \{h\} has the education in \{t\}.\\
        & currentMember & \{t\} has been an active member of \{h\}. \\
        & product & \{h\} has the product named \{t\}. \\
        & birthPlace & \{t\} is the birth place of \{h\}.\\
        & managerClub & \{h\} has been the manager of the club named \{t\}. \\
        & thirdTeam & \{t\} has been the third team in \{h\}. \\
        & writer & \{t\} is the writer of \{h\}. \\
        & president & \{t\} was the president when \{h\} was in office.\\
        & city & \{h\} is in the city of \{t\}. \\
        \midrule
        \multirow{15}{*}{\shortstack{YAGO310 \\ \\ (37 total \\ relation types)}} & isLocatedIn & \{h\} is located in \{t\}. \\
        & playsFor & \{h\} has played for \{t\}. \\
        & isAffiliatedTo & \{h\} has been affiliated to \{t\}. \\
        & diedIn & \{h\} died in \{t\}. \\
        & actedIn & \{h\} acted in \{t\}. \\
        & graduatedFrom & \{h\} graduated from \{t\}. \\
        & wasBornIn & \{h\} was born in \{t\}. \\
        & hasGender & \{t\} is the gender of \{h\}. \\
        & happenedIn & \{h\} happened in \{t\}. \\
        & hasMusicalRole & \{t\} is the musical role of \{h\}. \\
        & isConnectedTo & \{h\} is connected to \{t\}. \\
        & isMarriedTo & \{h\} is married to \{t\}. \\
        & participatedIn & \{h\} participated in \{t\}. \\
        & hasOfficialLanguage & \{t\} is an official language of \{h\}. \\
        & hasWonPrize & \{h\} has won the \{t\} prize. \\
        \bottomrule    
    \end{tabular}}
    \caption{\textbf{Exemplary templates to construct a question prompt for each relation type.}}
    \label{tab:templates}
\end{table*}

\begin{table*}[t]
    \centering
    \resizebox{0.9\textwidth}{!}{
    \begin{tabular}{cccccc}
         \toprule
         Dataset & Sample Percentage & \# Train Entities & \# Val Entities & \# Test Entities & \# Test - Train Entities \\
         \midrule
         DBP 100k & 0.01 & 8494 & 1218 & 1508 & 1127 \\
         & 0.001 & 1421 & 232 & 347 & 191 \\
         \midrule
         YAGO310 & 0.01 &  12763 & 1973 & 2016 & 1358 \\
         & 0.001 & 1551 & 201 & 230 & 209 \\
         \bottomrule
    \end{tabular}}
    \caption{Statistics of the dataset splits.}
    \label{tab:data_stats}
\end{table*}

\section{Discussion on Theoretical Understanding}\label{app:extended_discussion}

Due to the nature of the task, it is possible that the same embeddings that estimate an LLM’s knowledge on one dataset may not necessarily estimate its knowledge on another dataset. We do not claim or recommend any one embedding model to estimate knowledge of an LLM for all datasets and our main focus is in presenting a systematic evaluation framework to test it. It is thus highly difficult to theoretically understand estimation given the lack of pre-training data for both embeddings and large language models. 
For completeness and to explore any such correlations from the information that we know, we first categorize the embeddings in Table~\ref{tab:emb_explain} based on what they encode and their training paradigm and data. 
This shows that modern embedding architectures are more powerful in estimating LLM-acquired knowledge, such as Mistral-based NVE2 and Linq and Qwen-based GTE. Furthermore, we also find that instruction-tuning and hard negative mining is generally a preferred way to train these models to estimate LLM knowledge. However, the exact performance gaps depend on what kind of negative mining is done to train these models. It can also thus, be argued that LLM-based negative mining would naturally lead to a better estimation of LLM knowledge. Furthermore, we can also note that specialized embedding models such as ULTRA for graph structures are not quite suitable to estimate how LLMs encode knowledge which leads to interesting questions about what/how they encode.



\begin{table}[t]
    \centering
    \begin{tabular}{c|ccc}
        \toprule 
        Dataset & LLM & Sample percentage & Accuracy \\
        \midrule
        YAGO310 & GPT-4o & 0.01 & 46.23 \\
        & GPT-4o & 0.001 & 44.69 \\
        & GPT-4o-mini & 0.01 & 46.43 \\
        & GPT-4o-mini & 0.001 & 43.11 \\
        & GPT-4o COT & 0.01 & 60.71 \\
        & GPT-4o COT & 0.001 & 56.30 \\
        & GPT-4o-mini COT & 0.001 & 50.88 \\
        & Llama-3.1-8B & 0.01 & 42.78 \\
        & Llama-3.1-8B & 0.001 & 63.94 \\
        & Llama-3.3-70B & 0.01 & 52.06 \\
        & Llama-3.3-70B & 0.001 & 52.06 \\
        \midrule
        DBP100k & GPT-4o & 0.01 & 39.69 \\
        & GPT-4o & 0.001 & 66.45 \\
        & GPT-4o-mini & 0.01 & 39.74 \\
        & GPT-4o-mini & 0.001 & 64.63 \\
        & GPT-4o COT & 0.01 & 59.47 \\
        & GPT-4o COT & 0.001 & 76.91 \\
        & GPT-4o-mini COT & 0.01 & 51.27 \\
        & GPT-4o-mini COT & 0.001 & 73.01 \\
        & Llama-3.1-8B & 0.01 & 45.35 \\
        & Llama-3.3-70B & 0.01 & 51.02 \\
        & Llama-3.3-70B & 0.001 & 51.02 \\
        \bottomrule
    \end{tabular}
    \caption{Accuracy of LLMs on the Binary Generation task assuming 0 negative samples.}
    \label{tab:base_accuracy}
\end{table}

\begin{table}[t]
    \centering
    \begin{tabular}{c c c }
        \toprule
        Embedding & DBP-100k & YAGO-310 \\
        \midrule
        NVE2 &  0.0917 & 0.0795 \\
        GIST &  0.1130 & 0.0851 \\
        Linq &  \textbf{0.0886} & \textbf{0.0680} \\
        MXBAI &  0.1123 & 0.0831 \\
        \bottomrule
    \end{tabular}
    \caption{\textbf{Binary activation prediction:} Mean absolute error in predicting the adapted logits from the 15th layer of Llama3.1-8B on $1\%$ data. Lesser is better.}
    \label{tab:binary_activation001}
\end{table}

\begin{table*}[t]
    \resizebox{1.0\textwidth}{!}{%
    \begin{tabular}{c c c c c}
        \toprule
        Embedding model & Encode & Architecture & Training data & Training paradigm \\
        \midrule
        MPNET & Sentence & BERT and XLNet & 1B sentence pairs & Contrastive learning \\
        NVE2 & Sentence & Mistral7B-v0.1 + latent-attention pooling & Unknown & Instruction-tuning and hard-negative mining \\
        Linq & Sentence & E5-mistral7b-instruct and Mistral7B-v0.1 & LLM-generated + benchmark & Instruction tuning + Teacher-guided negative mining \\
        GTE & Sentence & Qwen7B & Unknown & Unknown \\
        GIST & Sentence & BAAI/bge-large-en-v1.5 & MEDI + mined triplets from MTEB & Guided In-sample Selection of Training Negatives \\
        MXBAI & Sentence & Matryoshka + binary quantization & Unknown & Instruction + contrastive-tuning \\
        ULTRA & Knowledge Graph & NBFNet & FB15k-237, DBP100k, YAGO310, etc. & Self-supervised contrastive learning \\
        \bottomrule
    \end{tabular}
    }
    \caption{Embedding categorization for understanding knowledge estimation.}
    \label{tab:emb_explain}
\end{table*}

\begin{table*}[t]
    \centering
    \resizebox{1.0\textwidth}{!}{
    \begin{tabular}{c c c c c | c c c c c c c}
        \toprule
        Dataset & Metric & Majority & Random & Llama & 
        MPNET & NVE2 & GIST & Linq & GTE & MXBAI & ULTRA \\

        \midrule
        \multirow{2}{*}{DBP-100k} &  AUC $\uparrow$ & - & - & 67.01 & \textbf{51.14} & 51.07 & 51.23 & 47.51 & 49.79 & 49.45 & 49.45\\
        & ACC $\uparrow$ & 52.40 & 51.29 & 61.62 & \textbf{51.47} & 50.63 & 50.81 & 50.48 & 50.76 & 50.88 & 49.35 \\
        \midrule
  
        \multirow{2}{*}{YAGO-310} & AUC $\uparrow$ & - & - & 58.77 & 48.50 & 50.55 & \textbf{50.70} & 47.61 & 48.94 & 50.65 & 49.75\\
        & ACC $\uparrow$ & 50.72 & 50.45 & 52.89 & 51.14 & 51.06 & \textbf{51.23} & 47.51 & 49.79 & 49.45 & 49.45 \\
        \bottomrule
    \end{tabular}
    }
    \caption{\textbf{Binary Generation on Llama-3.3-70B:} Accuracy (ACC) and AUC for predicting LLM-generated binary truth for $0.1\%$ true facts sampled from knowledge graphs.  
    }
    \label{tab:binary_gen_llama70b}
\end{table*}

\begin{table*}[t]
    \centering
    \begin{tabular}{l}
        \toprule
        \multicolumn{1}{c}{\textbf{Correct estimation}} \\
        \midrule
        University of Edinburgh is an alma mater of Dixie Carter. \\
        Berklee College of Music is an alma mater of Hiro Kanagawa. \\
        Toyota Corolla (E170/E180) has been assembled in Durban. \\
        Mass Mental is an associated band of Stevie Salas. \\
        Kerala is the birth place of Sathaar. \\
        3rd Marine Division fought in the battle of Calhoun. \\ 
        New Hampshire is the broadcasting area of Bally Sports Midwest. \\
        Ibaraki is in the country of Japan. \\
        Alf Kjellin is the director of Adventure Time.\\
        Luis Milla has been the manager of the club named England national under-21 association football team.\\
        \midrule
        \multicolumn{1}{c}{\textbf{Incorrect estimation}} \\
        \midrule
        Université catholique de Louvain has an affiliation with Coimbra Group. \\
        Sap is an associated band of The Greencards.\\
        Curve is an associated band of Joe Satriani.\\
        Canada is the birth place of Norman Jewison.\\
        Biddeford is a type of city.\\
        Rochester Americans is a former team of Bob Gracie.\\
        Peabody is a part of Essex County.\\
        Local Government Act 2000 is the name of a leader from Folkestone and Hythe.\\
        Sudan is the place of Mahdist War.\\
        George VI was the monarch when Walter Runciman, 1st Viscount Runciman of Doxford was in office.\\
        \bottomrule
    \end{tabular}
    \caption{Examples of correct and incorrect estimation by the Linq embedding for GPT-4o in DBP100k.}
    \label{tab:examples}
\end{table*}

\end{document}